\documentclass[letterpaper, 10 pt, conference]{ieeeconf}  

\usepackage{booktabs}
\usepackage{amsmath}
\usepackage{amssymb} 
\usepackage{color}
\usepackage{xcolor}
\usepackage{subcaption}
\usepackage{graphicx, siunitx}
\usepackage{soul}
\usepackage[hidelinks]{hyperref}
\usepackage[normalem]{ulem}
\usepackage{adjustbox}
\newcommand{\etal}{\textit{et al}. }

\IEEEoverridecommandlockouts                              

\title{\LARGE{\bf BEVal: A Cross-dataset Evaluation Study of \\ BEV Segmentation Models for Autonomous Driving} \\  \tiny{\phantom{spacing}} }

\author{Manuel Diaz-Zapata$^{1,2}$, Wenqian Liu$^{2}$, Robin Baruffa$^{2}$, Christian Laugier$^{2}$ 
\thanks{This work was partially supported by Toyota Motor Europe.}
\thanks{ 
$^{1}$CHROMA team, Univ Lyon, Inria, INSA Lyon, CITI Lab, France. $^{2}$CHROMA team, Univ. Grenoble Alpes, Inria, Grenoble, France. 
Correspondence: {\tt\small manuel.diaz-zapata@inria.fr}}%
}
\begin{document}

\maketitle
\thispagestyle{empty}
\pagestyle{empty}

\begin{abstract}

Current research in semantic bird's-eye view segmentation for autonomous driving focuses solely on optimizing neural network models using a single dataset, typically nuScenes. This practice leads to the development of highly specialized models that may fail when faced with different environments or sensor setups, a problem known as domain shift. In this paper, we conduct a comprehensive cross-dataset evaluation of state-of-the-art BEV segmentation models to assess their performance across different training and testing datasets and setups, as well as different semantic categories. We investigate the influence of different sensors, such as cameras and LiDAR, on the models' ability to generalize to diverse conditions and scenarios. Additionally, we conduct multi-dataset training experiments that improve models' BEV segmentation performance compared to single-dataset training. Our work addresses the gap in evaluating BEV segmentation models under cross-dataset validation. And our findings underscore the importance of enhancing model generalizability and adaptability to ensure more robust and reliable BEV segmentation approaches for autonomous driving applications. The code for this paper available at \color{blue}\texttt{\url{https://github.com/manueldiaz96/beval/}}\color{black}.
\end{abstract}

\section{Introduction}

Recently, the Bird's Eye View (BEV) representation has gained significant attention in the autonomous driving community as a crucial tool for scene understanding. Unlike traditional image or point cloud segmentation, the BEV encodes rich scene representations as a unified space for integrating information from multiple sensor modalities, offering advantages such as object size invariance and reduced occlusions \cite{robotics-book}. Inspired by the Binary Occupancy Grids \cite{Elfes-1989-OccGrid}, recent methodologies aim to develop a BEV representation enriched with semantic information encoded in each cell.

\begin{figure}[t]
    \centering\includegraphics[width=\linewidth]{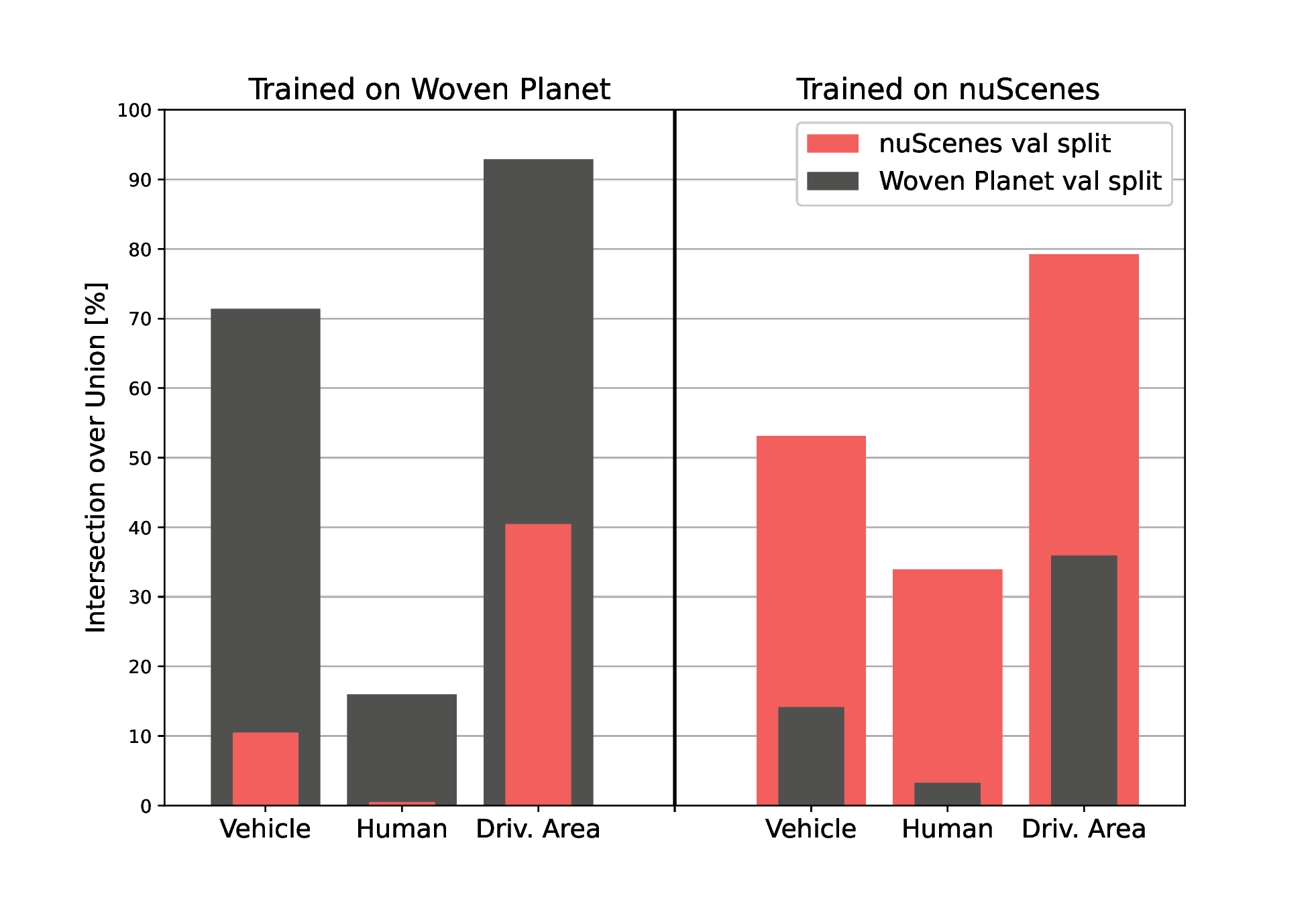}
\caption{Cross-dataset validation using the BEV semantic segmentation model LAPT-PP \cite{diaz2023laptnet}. IoU scores for three semantic categories are shown, with models trained on Woven Planet (left) and nuScenes (right). Pink bars indicate testing on nuScenes, gray on Woven Planet. Significant performance drops occur when training and testing datasets differ, revealing limited generalization.}
\label{fig:bar_graph}
\vspace{-5mm}
\end{figure}

Creating semantic BEV grid representations poses unique challenges, with approaches leveraging camera-based geometry \cite{philion-lss,imgs2map}, 3D point clouds \cite{pillarsegnet}, and more recently, sensor fusion techniques \cite{harley-bev, diaz2023laptnet, transfusegrid}. These top-down perspectives, useful for tracking \cite{Dequaire-2017-IJRR} and planning \cite{philion-lss}, enable systems to distinguish between various object types and scene areas. However, such models usually require large volumes of diverse, accurately annotated data \cite{udl-book}. Current BEV segmentation research predominantly uses the nuScenes dataset for both training and evaluation, raising concerns about model robustness and generalizability. While domain adaptation techniques can improve generalization, they often introduce complexity. Cross-dataset evaluation offers a more direct, empirical verification of model robustness across different real-world conditions without additional training or fine-tuning.

Moreover, cross-dataset evaluation contributes to establishing standardized benchmarks, reveals inherent model limitations and strengths, offering clear insights into how it will perform when deployed in varied real-world settings. Despite the importance and advantages of cross-dataset validation, which ensures that models generalize well beyond their training data and mitigates overfitting, this area remains underexplored in the BEV semantic grid segmentation literature. 

In this work, we aim to address this gap in evaluating BEV segmentation models across multiple datasets to verify their reliability and applicability in diverse real-world scenarios. We propose a novel cross-dataset framework for training and evaluating three BEV segmentation models on the nuScenes \cite{nuscenes} and Woven Planet datasets \cite{wovendataset}. 
We conduct experiments on three state-of-the-art BEV semantic segmentation models, evaluating their performance across three semantic categories using the Intersection Over Union (IoU) score. As shown in Fig. \ref{fig:bar_graph}, there is a significant performance drop when models are tested on unseen data from a different dataset. Our proposed cross-dataset validation framework aims to identify specific weaknesses or failure modes that may not be apparent within a single dataset, helping to develop more robust and reliable models for autonomous driving. To our knowledge, this is the first work to address such topic.

Our contributions are:
\begin{itemize}
\item We introduce the first cross-dataset validation framework for BEV semantic segmentation task. This framework is flexible, which can be extended to additional models, datasets and semantic categories.
\item We perform a comparative study using two real-world large-scale datasets, assessing three BEV segmentation models with a variety of input sensor modalities, across three semantic segmentation categories.
\item Additionally, we investigate the models' generalization ability by training them simultaneously on both datasets.
\end{itemize}

\section{Related work}

\subsection{Semantic BEV Segmentation}

Recent works in BEV semantic segmentation for autonomous driving include LiDAR-based approaches \cite{pillarsegnet}, learned image-to-BEV projection techniques \cite{philion-lss, fiery}, and attention-based methods using RGB images \cite{imgs2map, cross-view-transformers, bevformer}. Sensor fusion approaches combine camera and LiDAR \cite{diaz2023laptnet, transfusegrid, harley-bev}, as well as radar \cite{harley-bev} data to leverage their complementary strengths. While these diverse approaches give impressive results on BEV segmentation, most are primarily trained and evaluated on the nuScenes dataset, raising concerns about their robustness and generalizability. To address this, we introduce our cross-dataset validation for BEV segmentation models, aiming to develop more reliable and adaptable systems.

\begin{figure*}[t]
  \centering
  
  \begin{subfigure}[t]{.25\textwidth}
    \centering\includegraphics[width=\linewidth]{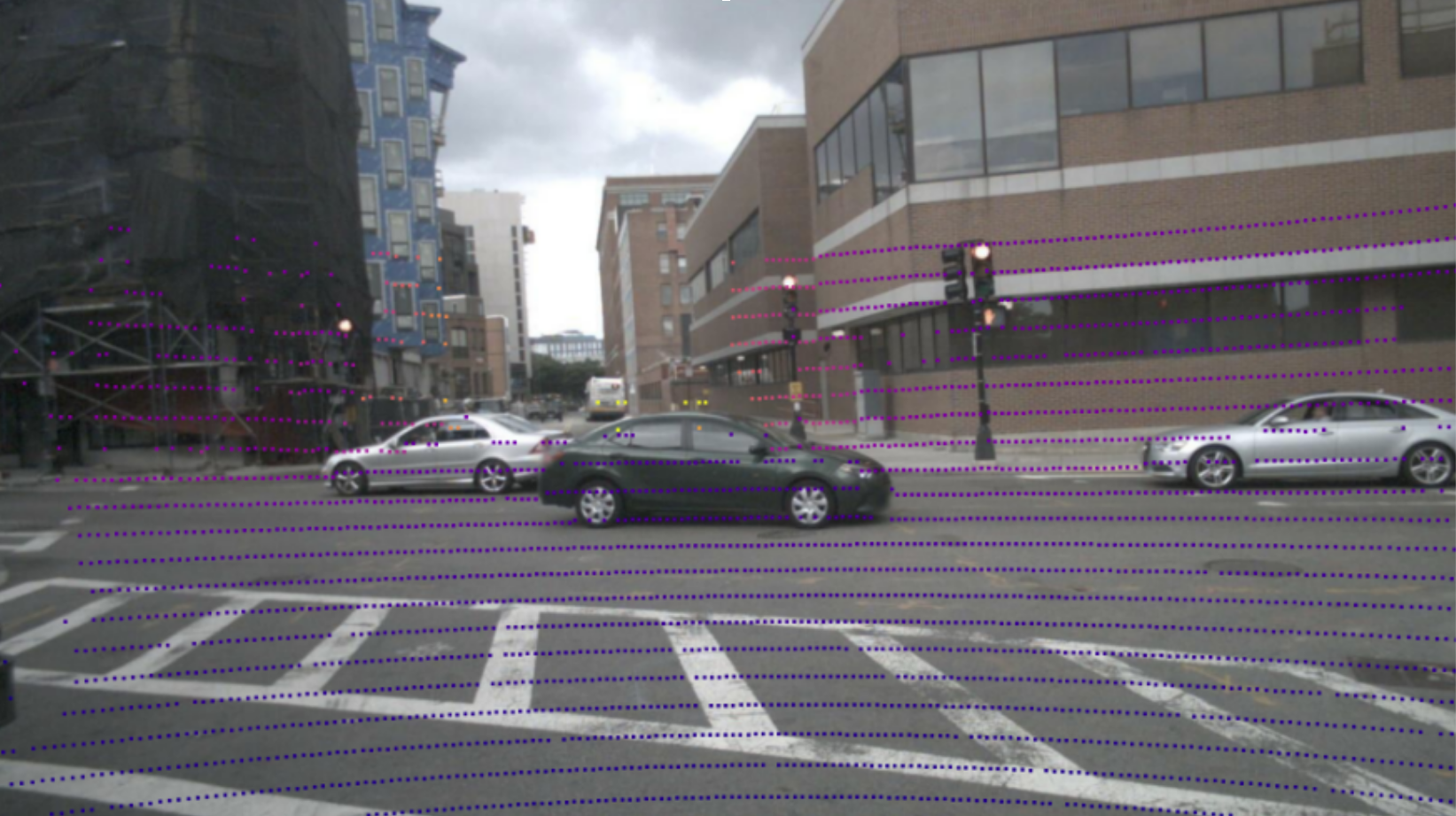}
    \centering\includegraphics[width=\linewidth]{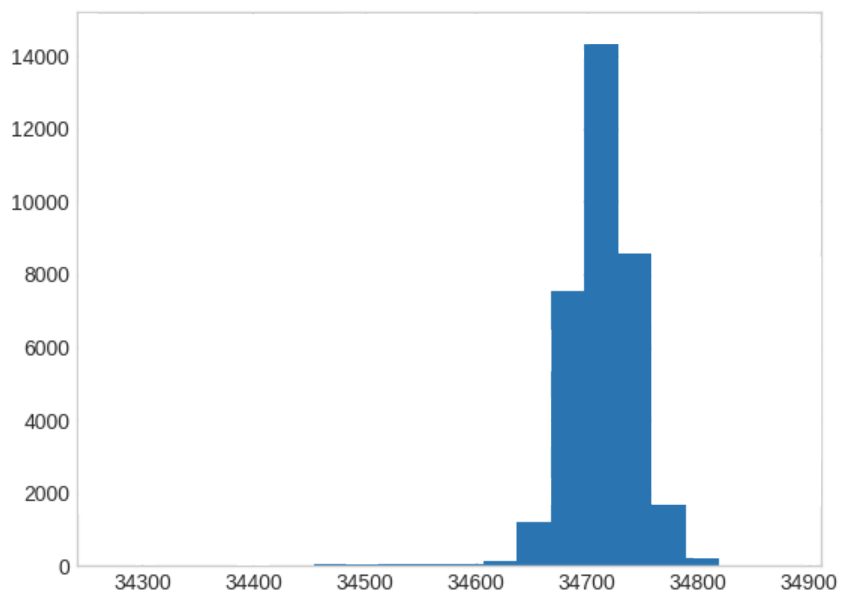}
    \centering\caption{nuScenes}
    \label{fig:histograms-nusc}
  \end{subfigure}
  \begin{subfigure}[t]{.25\textwidth}
    \centering\includegraphics[width=\linewidth]{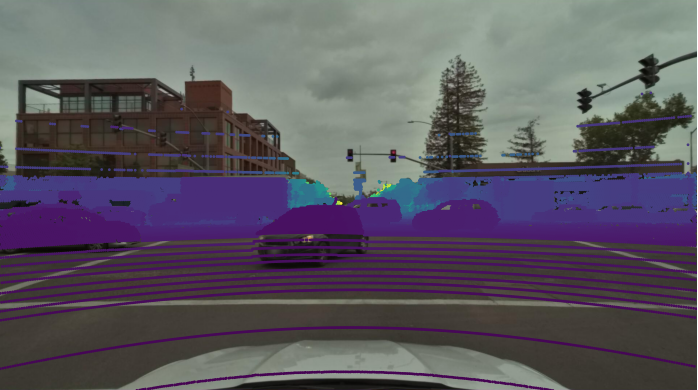}
    \centering\includegraphics[width=\linewidth]{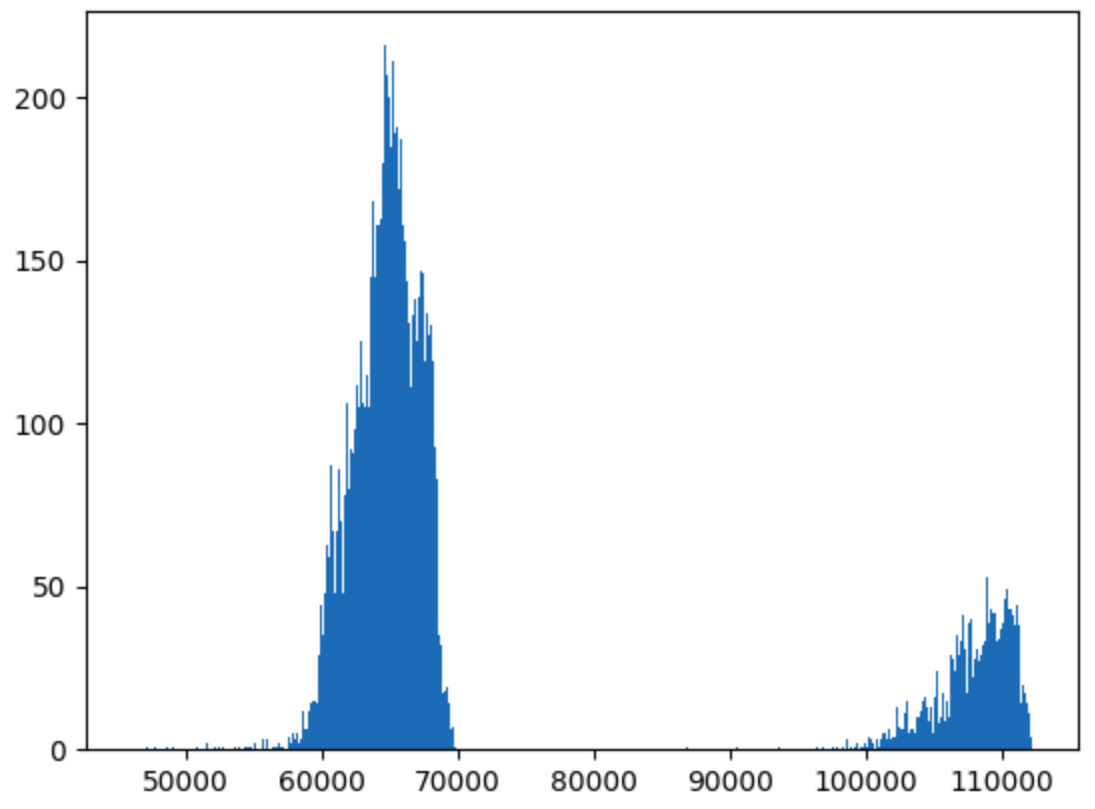}
    \caption{Woven Planet}
    \label{fig:histograms-lyft}
  \end{subfigure}
  \begin{subfigure}[t]{.25\textwidth}
    \centering\includegraphics[width=\linewidth]{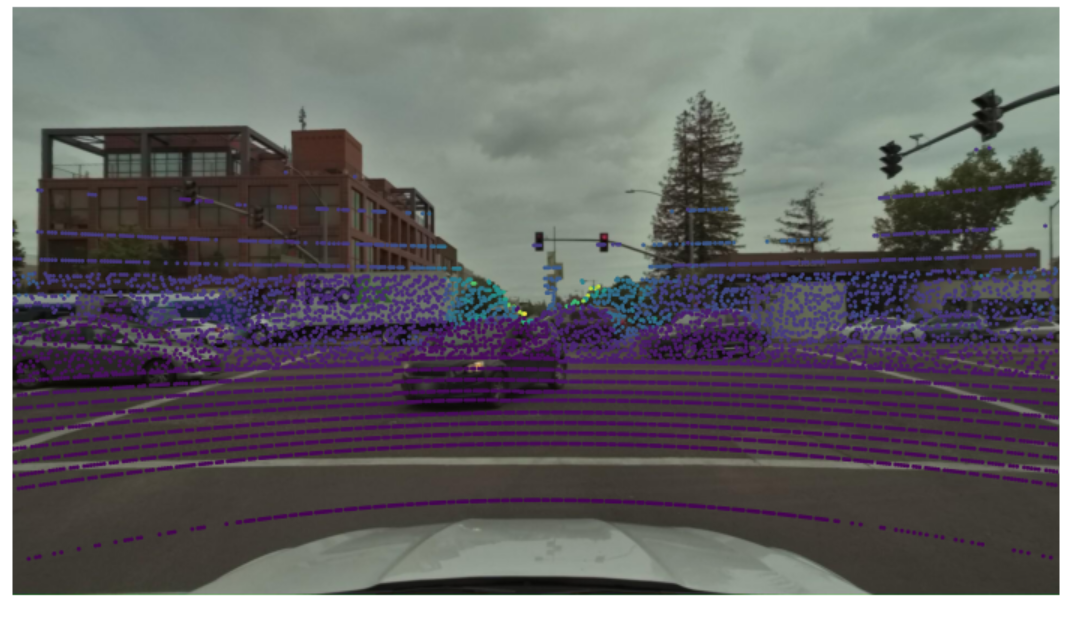}
    \centering\includegraphics[width=\linewidth]{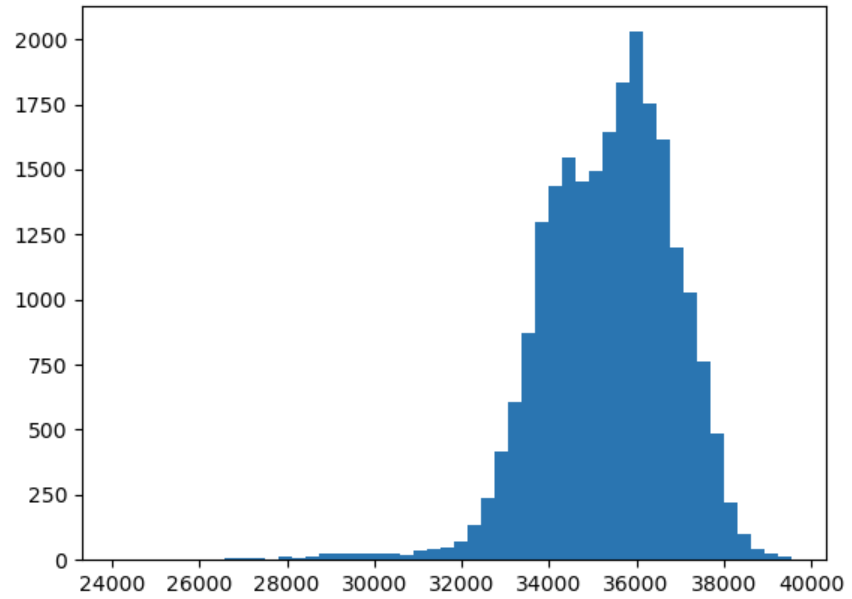}
    \caption{Woven Planet (subsampled)}
    \label{fig:histograms-lyft-subsampled}
  \end{subfigure}
\caption{Point cloud sample illustration (top) and histogram of the number of points per sample (bottom) for (a) nuScenes, (b) Woven Planet and (c) the subsampled Woven Planet point clouds. Best viewed with digital zoom.}
\label{fig:histograms}
\vspace{-5mm}
\end{figure*}



\subsection{Cross-dataset Validation}

In the field of autonomous driving there has been an interest in cross-dataset validation across different applications. Gilles \etal \cite{gilles2022corr} evaluated vehicle trajectory
prediction methods across four datasets, highlighting that the size of the dataset is not the most contributing factor in increasing performance, but rather its ability to faithfully represent real conditions. Gesnouin \etal \cite{gesnouin2022assessing}, studied pedestrian intention prediction across three datasets, finding that models often overfit to one dataset and underperform on others, highlighting the need for quantifying a model's uncertainty when evaluating on unseen data. St\"acker \etal \cite{stacker2023cross} evaluated a 3D detection network across two datasets using a camera and radar fusion approach,  demonstrating that visual variability in pretraining benefits camera features but not radar features, while the fusion of both modalities leads to the best performance overall. Furthermore, they only evaluated their models on the same dataset used for training. Despite the exploration of cross-dataset validation in autonomous driving tasks, there is a notable lack of research in BEV semantic grid segmentation, a gap that our work aims to fill.

\section{Methodology}

In this section, we outline the methodology of our study. First, we discuss the datasets utilized, highlighting their characteristics and differences. Next, we detail the processing of sensor data common to both datasets and the generation of ground truth. Finally, we describe the models, as well as the training and evaluation strategies employed in our study.

\subsection{Datasets}

To conduct cross-dataset evaluation study in this paper, we use nuScenes dataset \cite{nuscenes} and Woven Planet Perception Dataset \cite{wovendataset}, given their relevance in the BEV segmentation literature \cite{philion-lss, harley-bev, fiery} and their comparable sensor setups. The \textbf{nuScenes dataset} \cite{nuscenes} focuses on driving-specific scenarios and was collected in Boston (USA) and Singapore. It provides sensor information from six cameras, five radars and one 32-layer LiDAR across 1000 driving scenes of 20 seconds each. Given the different sampling rates of each sensor, the dataset provides a set of synchronized keyframes across all sensors with a frequency of 2Hz. It also provides 3D bounding box annotations for the different agents in each scene and a set of high-definition maps of the traversed areas. The \textbf{Woven Planet Perception Dataset} \cite{wovendataset}, formerly known as the Lyft Level 5 Perception Dataset, is a large-scale dataset for research on self-driving vehicles. Captured across the city of Palo Alto (USA), it provides sensor data from a set of six cameras and three 64-layer LiDARs, 3D bounding box annotations for pedestrians and vehicles in the scene, as well as a semantic map raster at a resolution of 10cm/px.

Both of the two datasets provide a comprehensive 360$^\circ$ field of view, including six surrounding cameras and a roof-mounted LiDAR. To maintain consistency in sensor configurations between both datasets, we omitted the use of radar sensors from the nuScenes dataset and the point clouds from the frontal LiDARs from the Woven Planet dataset.

\subsection{Point Cloud Processing}

Given the difference in the LiDAR specifications between nuScenes (32 layers) and Woven Planet (64 layers), we conducted a preliminary study to compare the distribution of point clouds across both datasets. We generated histograms depicting the number of points per sample, as shown in Fig. \ref{fig:histograms-nusc} for nuScenes and Fig. \ref{fig:histograms-lyft} for Woven Planet. A considerable difference in point cloud density was observed between the two datasets, which is expected due to their different LiDAR systems. Notably, nuScenes' point clouds exhibit greater uniformity across samples, with both the median and average number of points being 34,720, whereas Woven Planet's point clouds have an average of 72,431 points and a median of 65,568 points. 

To achieve a uniform number of points per sample across both datasets, we subsampled the point clouds in the Woven Planet dataset to match, as closely as possible, the density of those in the nuScenes dataset. We first transformed each point cloud available in the Woven Planet from the original Cartesian coordinates $(x,y,z)$ to spherical coordinates $(\rho, \theta, \phi)$. Then we divided the range of $\theta$ values into 32 sectors, corresponding to the 32 LiDAR layers in nuScenes. For the $\phi$ values, we divided them into 1500 sectors, as this produced a distribution most similar to nuScenes. We sampled one point from each sector and saved the resulting point cloud for later use. The difference between the original and subsampled point clouds is shown in the top of Fig. \ref{fig:histograms-lyft} and Fig. \ref{fig:histograms-lyft-subsampled} respectively. The histogram at the bottom of Fig. \ref{fig:histograms}c illustrates that the subsampled point cloud distribution in Woven Planet is finally closer to the nuScenes distribution, with a median of 35,498 points and a mean of 35,360 points.

\subsection{Image Processing}

The nuScenes and the Woven Planet datasets present different image sizes for their camera input. nuScenes provides a set of six camera images of $(1600 \times 900)$ pixels. In contrast, Woven Planet includes two different image sizes across scenes: some are $(1920 \times 1080)$ pixels with a $16:9$ aspect ratio (same as nuScenes), and other scenes are $(1124 \times 1024)$ pixels with a $1:1.1$ aspect ratio.

To ensure consistency across both datasets, we followed previous works \cite{philion-lss, diaz2023laptnet}, resizing and center cropping each image to have dimensions of $(128 \times 352)$ pixels. We also adjusted the intrinsic camera matrices accordingly. Additionally, we applied standard ImageNet \cite{deng2009imagenet} normalization before passing the images to the models for evaluation.

\subsection{Ground Truth Generation}

We evaluated BEV segmentation within a 100m by 100m area  surrounding the ego vehicle.  Consistent with previous studies \cite{philion-lss, fiery, harley-bev, diaz2023laptnet}, we discretized this space at a resolution of 0.5m per pixel, resulting in a 200 by 200 pixel grid. We used three semantic categories in our experiments, including the Human, Vehicle and the Drivable Area, given that these are the only semantic categories in common across both datasets. 

To obtain the required ground truth for Human and Vehicle classes, we discretized the provided 3D bounding box coordinates and sizes and projected them onto the BEV to generate the corresponding semantic ground truth. We did not filter these annotations based on visibility levels, as suggested in \cite{cross-view-transformers} and \cite{bartoccioni2022lara}, since these visibility levels are only available in the nuScenes dataset.

We generate the drivable area ground truth for the nuScenes dataset using its provided map API. By providing the ego position, area of interest and grid resolution, we can obtain a rasterization of the required map layer for any sample. For the Woven Planet dataset, which provides its map as one RGB image, the procedure for ground truth generation differs. First, we crop the area of interest from the original map image. Next, we apply a color filter to isolate pixels representing drivable areas and crossings, followed by a morphological closing operation using a $(5\times5)$ kernel to fill gaps left by the centerlines present in the map. Finally, we resize the image to match the required BEV resolution.

\subsection{Models}
\label{sec:models}

We perform cross-dataset validation experiments using state-of-the-art BEV semantic segmentation models with various input sensor modalities:

\noindent \textbf{Camera-only}: Lift-Splat-Shoot (LSS) \cite{philion-lss} is a prominent semantic BEV segmentation model that exclusively uses camera images as input. It predicts an implicit depth distribution to project image features into 3D space and assigns these features to BEV cells via sum-pooling. This model serves as a benchmark to assess performance variations across datasets when using only image inputs.

\noindent \textbf{Early Camera-LiDAR Sensor Fusion}: LAPT \cite{diaz2023laptnet} adopts an early fusion approach by combining camera and LiDAR data at the initial stage. This model utilizes LiDAR depth information to link image features with the BEV, projecting features from multiple image scales to enhance BEV coverage. It illustrates the impact of limited sensor fusion, focusing on depth values from point clouds rather than their complete 3D structure.

\noindent \textbf{Late Camera-LiDAR Sensor Fusion}: LAPT-PP \cite{diaz2023laptnet} is a variant of LAPT that employs late fusion techniques. This model integrates a LiDAR-specific encoder to generate BEV features exclusively from point-cloud data. The resulting feature map is then fused with camera-derived BEV features to predict final semantic segmentation. This model evaluates performance changes in networks relying on the 3D structure of point clouds, showcasing the effects of late-stage sensor fusion.

\begin{table*}[!htb]
    \begin{subtable}{0.5\textwidth}
    \centering
        \begin{tabular}{@{}lcc|cc@{}}
            \toprule
                                 & \textbf{NS} & \textbf{NS*} & \textbf{WP} & \textbf{WP*} \\ \midrule
            \textbf{LSS}         & 32.95       & 10.5   \color{gray}(68.13\% $\downarrow$)\color{black}      & 27.07       & 22.63   \color{gray}(16.4\% $\downarrow$)\color{black}     \\
            \textbf{LAPT}    & 47.03       & 22.3   \color{gray}(52.58\% $\downarrow$)\color{black}      & 57.98       & 37.18   \color{gray}(35.87\% $\downarrow$)\color{black}     \\
            \textbf{LAPT-PP} & 53.1        & 10.46  \color{gray}(80.3\% $\downarrow$)\color{black}      & 71.37       & 14.1    \color{gray}(80.24\% $\downarrow$)\color{black}     \\ \bottomrule
        \end{tabular}
        \caption{Vehicle}
    \end{subtable}
    \begin{subtable}{0.5\textwidth}
    \centering
        \begin{tabular}{@{}lcc|cc@{}}
            \toprule
                                               & \textbf{NS} & \textbf{NS*} & \textbf{WP} & \textbf{WP*} \\ \midrule
            \textbf{LSS}         & 12.21       & 0.4  \color{gray}(96.72\% $\downarrow$)\color{black}        & 5.55        & 4.1  \color{gray}(26.13\% $\downarrow$)\color{black}         \\
            \textbf{LAPT}    & 22.69       & 2.8  \color{gray}(87.66\% $\downarrow$)\color{black}         & 10.35       & 7.73  \color{gray}(25.31\% $\downarrow$)\color{black}        \\
            \textbf{LAPT-PP} & 33.95        & 0.5 \color{gray}(98.53\% $\downarrow$)\color{black}        & 15.95       & 3.25   \color{gray}(79.62\% $\downarrow$)\color{black}       \\ \bottomrule
        \end{tabular}
        \caption{Human}
    \end{subtable}
    
    \bigskip
    \centering
    \begin{subtable}{\textwidth}
    \centering
        \begin{tabular}{@{}lcc|cc@{}}
            \toprule
                                 & \textbf{NS} & \textbf{NS*} & \textbf{WP} & \textbf{WP*} \\ \midrule
            \textbf{LSS}         & 75.41       & 33.5   \color{gray}(55.58\% $\downarrow$)      & 84.88       & 48.55 \color{gray}(42.86\% $\downarrow$)       \\
            \textbf{LAPT}    & 77.15       & 47.6  \color{gray}(38.3\% $\downarrow$)       & 90.71       & 58.57 \color{gray}(35.43\% $\downarrow$)       \\
            \textbf{LAPT-PP} & 79.23       & 40.42  \color{gray}(48.98\% $\downarrow$)      & 92.88       & 35.9  \color{gray}(61.34\% $\downarrow$)       \\ \bottomrule
        \end{tabular}
        \caption{Drivable Area}
    \end{subtable}
    
    \bigskip
    \centering
    \begin{subtable}{\textwidth}
    \centering
        \begin{tabular}{@{}lcc|cc@{}}
            \toprule
                                 & \textbf{NS}   & \textbf{NS*}  & \textbf{WP}   & \textbf{WP*}  \\ \midrule
            \textbf{LSS}         & 19.4 / 62.03  & 12.77 \color{gray}(34.18\% $\downarrow$)\color{black}\, / 32.75 \color{gray}(47.2\% $\downarrow$) & 27.82 / 73.33 & 16.73 \color{gray}(39.86\% $\downarrow$)\color{black}\,  / 44.52 \color{gray}(39.29\% $\downarrow$)\\
            \textbf{LAPT}    & 15.5 / 60.0   & 19.13 \color{gray}(31.93\% $\uparrow$)\color{black}\,  / 36.11 \color{gray}(39.82\% $\downarrow$) & 23.62 / 68.07 & 16.05 \color{gray}(32.05\% $\downarrow$)\color{black}\,  / 49.28 \color{gray}(27.60\% $\downarrow$) \\
            \textbf{LAPT-PP} & 20.88 / 58.76 & 15.51 \color{gray}(24.72\% $\downarrow$)\color{black}\, / 37.04 \color{gray}(36.96\% $\downarrow$) & 43.3 / 68.33  & 9.08 \color{gray}(79.03\% $\downarrow$)\color{black}\, / 34.04 \color{gray}(50.18\% $\downarrow$) \\ \bottomrule
        \end{tabular}
        \caption{Vehicle / Drivable Area}
    \end{subtable}
    
    \caption{IoU [\%] scores for cross-dataset evaluation of single-class BEV segmentation: (a) Vehicle, (b) Human, (c) Drivable Area, and multi-class BEV segmentation with joint prediction: (d) Vehicle / Drivable Area. The columns with a asterisk sign ($\ast$) indicate models trained on one dataset and tested on the other separately. The values in gray beside the IoU scores denote the performance difference in percentage when models were trained on a different dataset compared to their baselines (trained and tested on the same dataset). And the arrows indicate whether the performance drops ($\downarrow$) or increases ($\uparrow$). (\textbf{NS} - nuScenes / \textbf{WP} - Woven Planet)}
\label{tab:1}
\vspace{-5mm}
\end{table*}

\subsection{Experimental Details}

We evaluated each of the models described in Section \ref{sec:models} for single-class BEV segmentation on Vehicle, Human, and Drivable Area respectively. Additionally, we also evaluated each model on multi-class BEV segmentation by jointly predicting Vehicle and Drivable Area, to assess their performance in a broader context.

For both single-class and multi-class segmentation tasks, each model underwent two distinct training setups. Firstly, models were individually trained on the nuScenes dataset and the Woven Planet dataset separately, then evaluated on both datasets to measure cross-dataset generalization. Secondly, models were trained simultaneously on both datasets to investigate the impact of augmented training data on performance across individual datasets. 

For training, we utilized binary cross-entropy loss for single-class prediction, and cross-entropy loss for multi-class prediction. Training employed a batch size of 10 and the Adam optimizer with a learning rate of $0.001$, continuing for 50 epochs or until convergence of the evaluation metric, whichever happens first. 

During evaluation, we adhered to standard practices in the field, using the Intersection over Union (IoU) metric to gauge the overlap between model predictions and ground truth annotations.

\section{Results and Discussion}


In this section, we first present the experimental results for cross-dataset evaluation for both single-class and multi-class BEV segmentation. Then we discuss the experiments on multi-dataset training setups. We will show results quantitatively and qualitatively. 

\subsection{Cross-dataset Evaluation}

We present quantitative results for cross-dataset evaluation in Table \ref{tab:1}. For single-class prediction results shown in Table \ref{tab:1}(a), \ref{tab:1}(b) and \ref{tab:1}(c), models trained on Woven Planet mostly exhibit worse generalization to nuScenes across different semantic classes compared to the reverse case, as indicated by the significant drop in IoU scores. 

Among the models, we observe that LAPT-PP suffers the most than the other two models in cross-dataset evaluation, showing the largest performance drops across different semantic classes and respective train-test setups. This significant degradation can be attributed to LAPT-PP-FPN's heavy reliance on LiDAR point cloud features. Models based on LiDAR often struggle with cross-dataset generalization due to sensor-specific variations, particularly the differences in LiDAR configurations between Woven Planet and nuScenes datasets. In contrast, image-based models, such as LSS and LAPT, benefit from more consistent visual data, standardized preprocessing, and annotation practices, resulting in better cross-dataset performance.

We also present multi-class segmentation results in Table \ref{tab:1}(d) by training and testing models to jointly predict Vehicle and Drivable Area in the BEV grid. We notice that multi-class prediction demonstrates less performance drop compared to single-class across datasets and models. This finding suggests that jointly predicting multiple classes (vehicle and drivable area) helps mitigate the impact of dataset variations on model performance. By jointly learning multi-class prediction, the model captures robust and redundant features from the scene that can help stabilize predictions against dataset-specific biases.

\begin{figure*}[t]
  \centering
  \begin{subfigure}{0.22\textwidth}
    \centering\includegraphics[width=\linewidth]{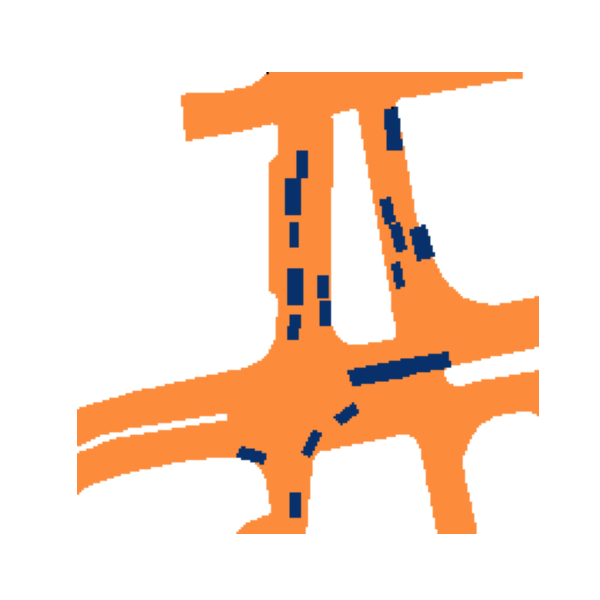}
    \centering\caption{Ground Truth}
    \label{fig:nusc-GT}
  \end{subfigure}
  \begin{subfigure}{0.22\textwidth}
    \centering\includegraphics[width=\linewidth]{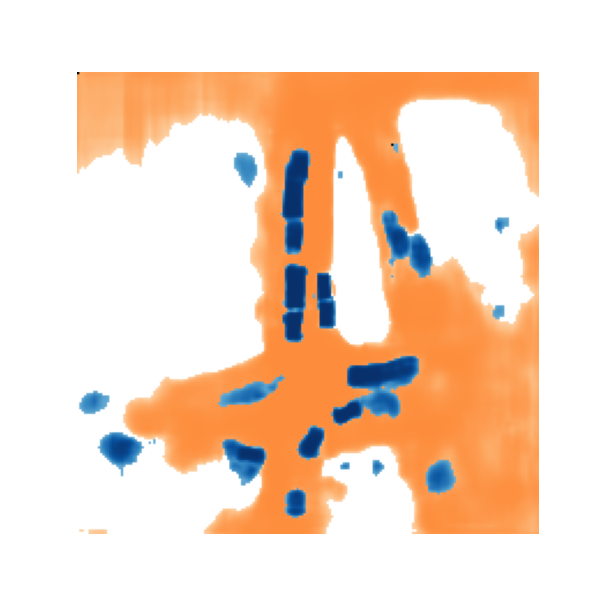}
    \centering\caption{Trained on Nuscenes}
    \label{fig:nusc-nusc}
  \end{subfigure}
  \begin{subfigure}{0.22\textwidth}
    \centering\includegraphics[width=\linewidth]{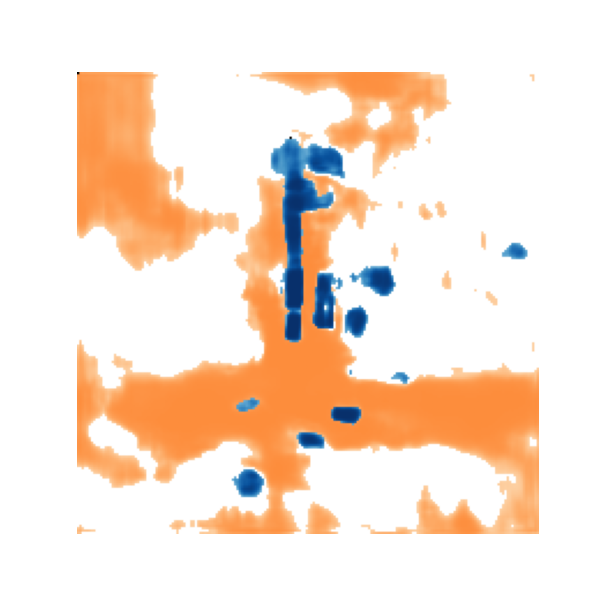}
    \centering\caption{\centering Trained on Woven Planet}
    \label{fig:nusc-lyft}
  \end{subfigure}
  \begin{subfigure}{0.22\textwidth}
    \centering\includegraphics[width=\linewidth]{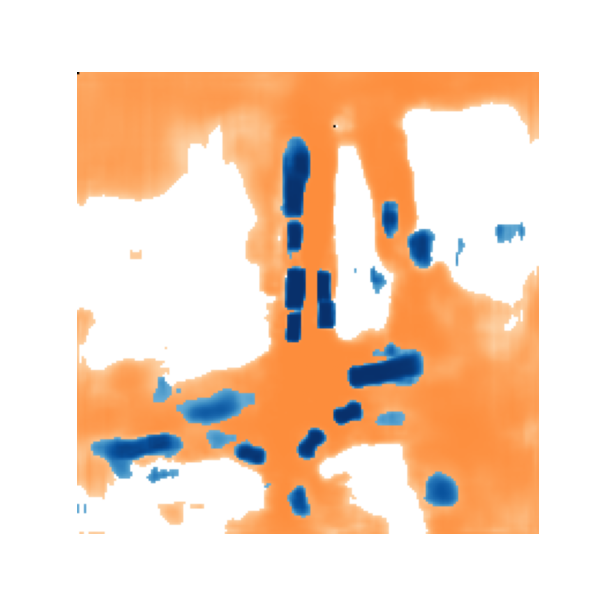}
    \centering\caption{Trained on both datasets}
    \label{fig:nusc-nusc-lyft}
  \end{subfigure}

  \caption{Qualitative BEV semantic segmentation results for LAPT-PP on nuScenes dataset.}
\label{fig:grids:1}

\end{figure*}

\begin{figure*}
 \centering
  \begin{subfigure}{0.22\textwidth}
    \centering\includegraphics[width=\linewidth]{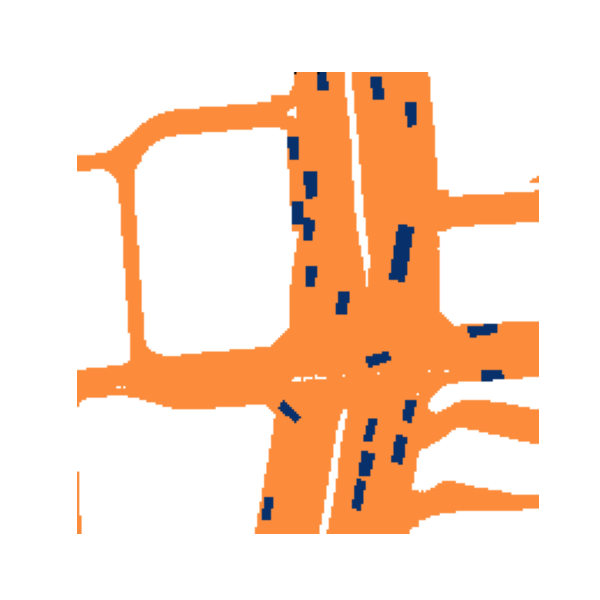}
    \centering\caption{Ground Truth}
    \label{fig:lyft-GT}
  \end{subfigure}
  \begin{subfigure}{0.22\textwidth}
    \centering\includegraphics[width=\linewidth]{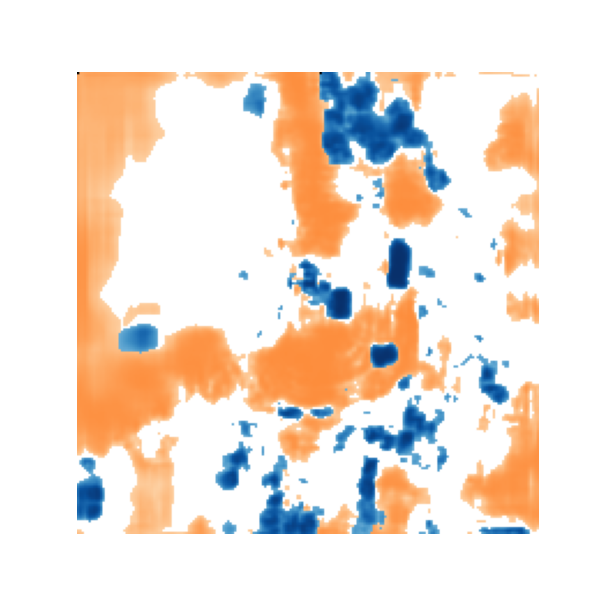}
    \centering\caption{Trained on Nuscenes}
    \label{fig:lyft-nusc}
  \end{subfigure}
  \begin{subfigure}{0.22\textwidth}
    \centering\includegraphics[width=\linewidth]{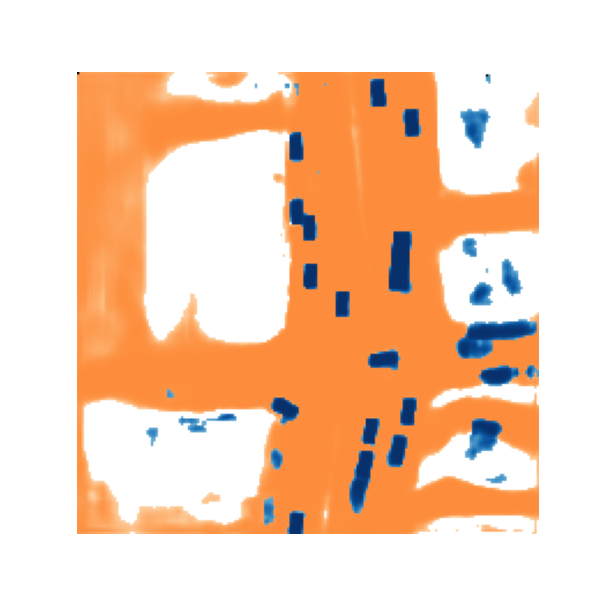}
    \centering\caption{Trained on Woven Planet}
    \label{fig:lyft-lyft}
  \end{subfigure}
  \begin{subfigure}{0.22\textwidth}
    \centering\includegraphics[width=\linewidth]{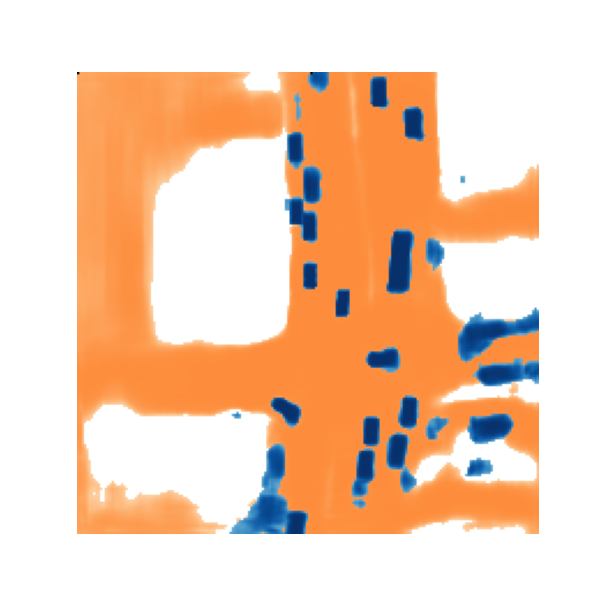}
    \centering\caption{Trained on both datasets}
    \label{fig:lyft-both}
  \end{subfigure}
  
\caption{Qualitative BEV semantic segmentation results for LAPT-PP on Woven Planet dataset.}
\label{fig:grids:2}

\end{figure*}

\begin{table*}[h]
    \centering
    \begin{tabular}{@{}lccccccccc@{}}
    \toprule
        \multicolumn{1}{c}{}          & \multicolumn{4}{c}{\textbf{Woven Planet}}                       &  & \multicolumn{4}{c}{\textbf{nuScenes}}                                         \\ \midrule
        \multicolumn{1}{c}{\textbf{}} & \textbf{Vehicle} & \textbf{Human} & \multicolumn{1}{c|}{\textbf{Driv. A.}} & \textbf{Vehicle / Driv. A.} &  & \textbf{Vehicle} & \textbf{Human} & \multicolumn{1}{c|}{\textbf{Driv. A.}} & \textbf{Vehicle / Driv. A.} \\ \cmidrule(lr){2-5} \cmidrule(l){7-10} 
        \textbf{LSS}             & 28.58 {\color{gray}($\uparrow$)} & 6.11 {\color{gray}($\uparrow$)} & \multicolumn{1}{c|}{83.19 {\color{gray}($\downarrow$)}}    & 25.95 {\color{gray}($\downarrow$)} / 70.22 {\color{gray}($\downarrow$)}    &  & 33.47 {\color{gray}($\uparrow$)}    & 12.35 {\color{gray}($\uparrow$)} &  \multicolumn{1}{c|}{76.63 {\color{gray}($\uparrow$)}}    & 24.74 {\color{gray}($\uparrow$)} / 65.52 {\color{gray}($\uparrow$)}     \\
        \textbf{LAPT}                   & 58.32 {\color{gray}($\uparrow$)}  & 11.96 {\color{gray}($\uparrow$)} & \multicolumn{1}{c|}{87.65 {\color{gray}($\downarrow$)}}    & 20.84 {\color{gray}($\downarrow$)} / 63.66 {\color{gray}($\downarrow$)}     &  & 48.55 {\color{gray}($\uparrow$)}  & 21.76 {\color{gray}($\downarrow$)}  & \multicolumn{1}{c|}{78.16 {\color{gray}($\uparrow$)}}    & 13.76 {\color{gray}($\downarrow$)} / 51.57 {\color{gray}($\downarrow$)}     \\
        \textbf{LAPT-PP}                & 70.57 {\color{gray}($\downarrow$)}  & 17.31 {\color{gray}($\uparrow$)} & \multicolumn{1}{c|}{90.57 {\color{gray}($\downarrow$)}}    & 29.75 {\color{gray}($\downarrow$)} / 66.03 {\color{gray}($\downarrow$)}   &  & 52.48 {\color{gray}($\downarrow$)}  & 33.76 {\color{gray}($\downarrow$)} & \multicolumn{1}{c|}{79.77 {\color{gray}($\uparrow$)}}    & 16.72 {\color{gray}($\downarrow$)} / 52.68 {\color{gray}($\downarrow$)}     \\ \bottomrule
    \end{tabular}
    \caption{Multi-dataset training for BEV semantic segmentation tasks. The models are trained on the combined training sets from both the Woven Planet and nuScenes datasets. They are then tested on each dataset separately. The IoU [\%] scores are calculated for single-class semantic segmentation: Vehicle, Human, Drivable Area, and multi-class prediction: Vehicle/Drivable Area. Additionally, the arrows in gray color indicate whether the performance drop ($\downarrow$) or increase ($\uparrow$) compared to the baseline scores shown in Table \ref{tab:1}. }
    \label{tab:2}
    \vspace{-5mm}
\end{table*}

Last but not least, we realize that the human segmentation performance yields the lowest absolute IoU scores across datasets and models. On one hand, human BEV segmentation is challenging given the limited number of samples available within datasets. On the other hand, the chosen BEV grid resolution of 0.5m/px results in each human annotation being represented by only one to two pixels. This fine granularity can lead to the development of highly specialized models when trained on individual datasets.

\subsection{Multi-dataset Training}

Next, we conducted experiments using the combined training sets from both the Woven Planet and nuScenes datasets. We then tested each model on each dataset separately and present the IoU scores in Table \ref{tab:2}. The performances for both single-class and multi-class segmentation are shown. After training on both datasets, the models demonstrated consistent accuracy across both datasets. Specifically, the IoU scores for each dataset were similar to their baselines shown in Table \ref{tab:1}. Additionally, the IoU scores across the two datasets exhibited balanced performance, without bias towards one dataset or the other. Notably, as nuScenes data is included in the training process, all models for Human segmentation improve by 10\% to 15\% on the Woven Planet. This improvement can be attributed to the more varied data in nuScenes, as noted by Gilles et al. \cite{gilles2022corr}, which helps the model better understand the task.

While both LSS and LAPT achieve improved performance on both datasets, LAPT-PP suffers a slight performance degradation, which is within our expectation. This is likely due to the domain shift between the two datasets, which have different data distributions (e.g., different sensor configurations, environmental conditions, annotation styles).

\subsection{Qualitative Results}

In Fig. \ref{fig:grids:1} and Fig. \ref{fig:grids:2}, we show qualitative results yielded by the LAPT-PP model on nuScenes and Woven Planet datasets respectively. We jointly predict Vehicle and Drivable Area in the scene onto the same BEV grid. Following the discussion above, the model performs the best on the dataset it was trained on. Fig. \ref{fig:grids:1}(b) and \ref{fig:grids:2}(c) show two examples when LAPT-PP was trained and tested using the same dataset, and the resultant BEV grids closely match the ground truth. However, when evaluating on a different dataset (Fig. \ref{fig:grids:1}(c) and Fig. \ref{fig:grids:1}(b)), the model's predictions fail to accurately represent the scene semantics. 

When the model is trained on both datasets, as shown in Fig. \ref{fig:grids:1}(d) and Fig. \ref{fig:grids:2}(d), it is able to represent most of the scene accurately but does not achieve the same level of performance as single-dataset training. For further qualitative model comparisons across datasets and classes, see the video: \color{red}\url{https://youtu.be/z9-wJ-FTc8Y}\color{black}

\section{Conclusions}

In this paper, we addressed the critical gap in cross-dataset evaluation research for BEV semantic grid segmentation tasks. We evaluated three BEV segmentation models across three semantic categories using two autonomous driving datasets. Additionally, we proposed a preprocessing procedure to standardize the setups of the two datasets, ensuring similar data distributions and groundtruth annotations. Our results indicate that models utilizing images as the primary feature source demonstrate superior generalization across datasets, whereas those relying on LiDAR point clouds are more sensitive to dataset-specific characteristics. Furthermore, our study suggests that multi-dataset training achieves performance comparable to single-dataset training, albeit with potential slight performance drops due to domain shift. These findings underscore the importance of diverse data exposure in developing robust and reliable autonomous driving systems.

In future work, we plan to explore other data augmentation methods and techniques in domain adaptation. These steps aim to identify specific factors contributing to performance drops and develop strategies to mitigate them, ultimately leading to models capable of generalizing to broader conditions and scenarios.

\section*{ACKNOWLEDGMENTS}
\small
The experiments presented in this paper were carried out using the Grid'5000 testbed, supported by a scientific interest group hosted by Inria and including CNRS, RENATER and several Universities as well as other organizations.

\bibliographystyle{IEEEtran}


\bibliography{egbib}   

\end{document}